\documentclass{article}

     \PassOptionsToPackage{numbers, compress}{natbib}

     \usepackage[preprint]{neurips_2020}

\usepackage{graphicx} 
\usepackage{subfigure}
\usepackage[utf8]{inputenc} 
\usepackage[T1]{fontenc}    
\usepackage{hyperref}       
\usepackage{url}            
\usepackage{booktabs}       
\usepackage{amsfonts}       
\usepackage{nicefrac}       
\usepackage{microtype}      
\usepackage{enumitem}
\newlist{steps}{enumerate}{1}
\setlist[steps, 1]{label = step \arabic*:}

\usepackage{verbatim}
\usepackage[noend]{algpseudocode}
\usepackage{algorithmicx,algorithm}

\title{Experience Augmentation: Boosting and Accelerating Off-Policy Multi-Agent Reinforcement Learning}

\author{%
  Zhenhui Ye, ~~Yining Chen, ~~Guanghua Song, ~~Bowei Yang, ~~Sheng Fan  \\
  School of Aeronatuics and Astronautics, Zhejiang University\\
  38 Zheda RD, Hangzhou, P.R. China\\
  \texttt{\{zhenhuiye, ch19930611, ghsong, boweiy, fansheng\}@zju.edu.cn} \\
}

\begin{document}

\maketitle

\begin{abstract}
  Exploration of the high-dimensional state action space is one of the biggest challenges in Reinforcement Learning (RL), especially in multi-agent domain. We present a novel technique called \emph{Experience Augmentation}, which enables a time-efficient and boosted learning based on a fast, fair and thorough exploration to the environment. It can be combined with arbitrary off-policy MARL algorithms and is applicable to either homogeneous or heterogeneous environments.
  
  We demonstrate our approach by combining it with MADDPG and verifing the performance in two homogeneous and one heterogeneous environments. In the best performing scenario, the MADDPG with experience augmentation reaches to the convergence reward of vanilla MADDPG with 1/4 realistic time, and its convergence beats the original model by a significant margin. Our ablation studies show that experience augmentation is a crucial ingredient which accelerates the training process and boosts the convergence.
\end{abstract}

\section{Introduction}
Reinforcement learning(RL) has achieved impressive results in challenging problems, from playing games \cite{teaching2014}\cite{mastering2016} to robotics\cite{end2016}\cite{path2017}. Nowadays, the biggest difficulty to deploy RL in real world is to collect large amounts of data to train a good RL policy for any new environment. \emph{Experience replay (ER)}  \citep{lin1992}, which is a fundamental component of off-policy reinforcement learning, partially address this problem by storing experience in a memory buffer and reusing them randomly for multiple times. It breaks the correlation between the streaming of training data and improve the data efficiency, which stabilizes the training process and leads to a better convergence result \citep{DQN2015}. Currently, a majority of off-policy reinforcement learning algorithms, such as DDPG \citep{DDPG2015}, C51 \citep{C51_2017}, SAC \citep{SAC2018},  have adopted experience replay for its performance and simplicity. 

However, in multi-agent domain, the problem is more complicated. Firstly, the MARL algorithms using \emph{centralized training framework} \citep{MADDPG2017} \citep{MFMARL2018}\citep{QMIX2020}, demand more exploration than those in single-agent domain, since the state action space expands exponentially as the number of agents grows, which is known as \emph{curse of dimensionality}. Secondly, the interaction between the RL agents and the environment, which is necessary to collect training data, is the most time-consuming part to train a reinforcement learning system \citep{PER2015} and its computational complexity grows rapidly as the the number of agents grows. These two factors make it a complicated but promising problem to make a fast exploration and train a multi-agent system in a time-efficient manner.

To improve the training efficiency, previous work have made improvements in several aspects of experience replay, such as importance sampling\citep{PER2015}, setting sub-goals to address sparse reward problem\citep{HER2017}, examining the effects of hyper-parameters\citep{CER2017}, sharing experiences among distributed agents\citep{DPER2018}, the utilization of on-policy experience \citep{mix_on_policy_2019} , etc. Our work, however, is based on an intrinsic nature of the multi-agent task, and can be combined with above techniques for further improvement.

In this paper, we introduce a technique called \emph{Experience Augmentation}, which provides a fast, thorough, unbiased exploration by shuffling the order of agents and accelerates the training by additionally updating the parameters on the generated experiences. Applied to MADDPG\citep{MADDPG2017}, a classical MARL baseline, the performance of experience augmentation is demonstrated in 3 scenarios. In the best performing scenario, the agent with experience augmentation technique achieves the convergence reward of vanilla MADDPG with 1/4 training time, and its convergence beats the original model by a significant margin.

\section{Preliminaries}
\label{preliminaries}
\paragraph{Experience Replay}
Before Deep Q-Learning\citep{DQN2015} demonstrated the effectiveness of \emph{Experience Replay} \citep{lin1992}, reinforcement learning was suffered from the instability mainly caused by the correlated data. To perform experience replay, it is typical to store the agent's experience $e=(o,a,r,o')$ at each step into the replay buffer $\cal{D}$ and uniformly sample $n$ mini-batches of experience to update the parameters for $n$ times in every $\tau$ steps. For ease of explanation, we refer $n$ to \emph{update times} and $\tau$ to \emph{training interval}. A large sliding window replay memory of size $B$ (such as 1 million) is usually used as the container of the stored experiences.

There are mainly three benefits of the experience replay mechanism. Firstly, by breaking the correlations in the stream of  training data, experience replay stabilizes the training and induces a better convergence result\citep{PER2015}. Secondly, when learning on-policy, the current parameters of the policy determine the next experience to be sampled, which is, however, the next training data to be used for updating the parameters. Under such a circumstance, a bad feedback is easy to occur and the parameters would frequently get stuck in a poor local minimum\citep{DQN2015}. By using experience replay, the distribution of training data is averaged over many of its previous states, ensuring the breadth of the data distribution and thus allows for a relatively thorough exploration and smooth learning process. Thirdly, experience replay also allows for less training time, as each step of experience is reused for several weight updates, it requires less interactions between agents and its environment, which is usually a time-consuming part in the learning process.
\paragraph{MADDPG}
As an extension of actor-critic policy gradient method in multi-agent domain, MADDPG was proposed to use the framework of \emph{centralized training with decentralized execution}, where the critic is augmented with extra information about the policies of other agents to help learn the policy effectively . In the execution phase, the actors would act in a decentralized manner, yet well-trained to cooperative or compete with other agents.

The centralized critic of agent $i$ is represented by $Q_i^\mu(\vec{o},\vec{a})$, where $\mu=\{\mu_1,\cdots,\mu_N\}$ is the collection of all agents' deterministic policy; $\vec{o}=(o_1,\cdots,o_N)$ and  $\vec{a}=(a_1,\cdots,a_N)$ are the concatenation of every agent's observation and action. Each agent $i$ trains its Q-function by minimizing the loss function \citep{q-learning_1992} :
\begin{equation}
\mathcal{L}=\frac{1}{S}\Sigma(y-Q^\mu_i(\vec{o},\vec{a}))^2
\end{equation}
where $y=r_i+\gamma Q_i^{\mu'}(\vec{o'},a'_1,\cdots,a'_N)|_{a'_k=\mu'_k(o'_k)}$ , and $S$ is the size of the mini-batch. 

The actor network of agent $i$ is trained by determinstic policy gradient\citep{DPG2014} to maximize the objective function $J$, whose gradient is given by:
\begin{equation}
\nabla_{\theta_i}J=\frac{1}{S}\Sigma\nabla_{\theta_i}\mu_i(o_i)\nabla_{\mu(o_i)}Q_i^\mu(\vec{o},a_1,\cdots,a_{i-1},\mu(o_i),a_{i+1},\cdots,a_N)
\end{equation}
In MADDPG, the update times $n$ is set to 1, and training interval $\tau$ is 100.

\section{Methodology}
\subsection{Two properties of the MARL environment}
\label{a motivating example}
As is fulfilled by a majority of multi-agent envrionments \cite{MADDPG2017}\cite{emergence2018}\cite{pursuit2002}\cite{cooperative2017}, we assume that the reward function for any agent $i$ can be represented as :
\begin{equation}
r_i=f_{g(i)}(o_i,a_i,\vec{o_{\verb|\|i}},\vec{a_{\verb|\|i}}) ~~~~~~\forall i =1,\cdots,N
\end{equation}
where $g(i)$ represents the group of agent $i$.

We also assume that each agent in a same group is homogeneous to others, i.e. they share the same properties(such as size, shape, mass, etc) and the same reward fuction. These assumptions lead to two properties of the environment:
\begin{enumerate}
	\label{property1}
	\item The reward function for agent $i$ is symmetric to any other agents by group, i.e., exchanging the observation and action between any other two agents in the same group, would not influence the work done by agent $i$ or change the result of the environmental step , making no difference in the reward of agent $i$.
	\label{property2}
	\item The reward information of any agent $j$ in the same group with agent $i$ can also be utilized to train agent $i$, i.e., exchanging the observation and action between agent $i$ and any agent $j$ in group $g(i)$, will exchange the reward as well.
\end{enumerate}

Given the two properties of the environment, we could augment the original experience by shuffling the order of agents in a specific way, as shown in Sec.\ref{ShuffleTrick}.

\subsection{Shuffle Trick}
\label{ShuffleTrick}
In Sec.\ref{a motivating example}, we provide two properties of the MARL environments, which are the intuitive idea of our method. In this section, we formally propose the technique called \emph{Shuffle Trick} that augments the original dataset factorially, and thus partially deteriorates the curse of dimensionality, which is caused by the exponentially expanded state action space of MARL.

There are two steps to perform the shuffle trick: firstly, find the feasible permutation matrix set $\cal{P}$ in which every permutation matrix shuffles every agent by group; secondly, randomly select a permutation matrix $P^k$ from $\cal{P}$ and premultiply it to the original experience to shuffle the agents' order. In short, the shuffle trick can be expressed as:
\begin{equation}
(\vec{o},\vec{a},\vec{r},\vec{o}')\longrightarrow(P^k\cdot \vec{o},P^k\cdot\vec{a},P^k\cdot\vec{r},P^k\cdot\vec{o}')~~~~~\forall{P^k}\in \cal{P}
\end{equation}

\begin{figure}
	
	\centering
	\includegraphics[width=8cm]{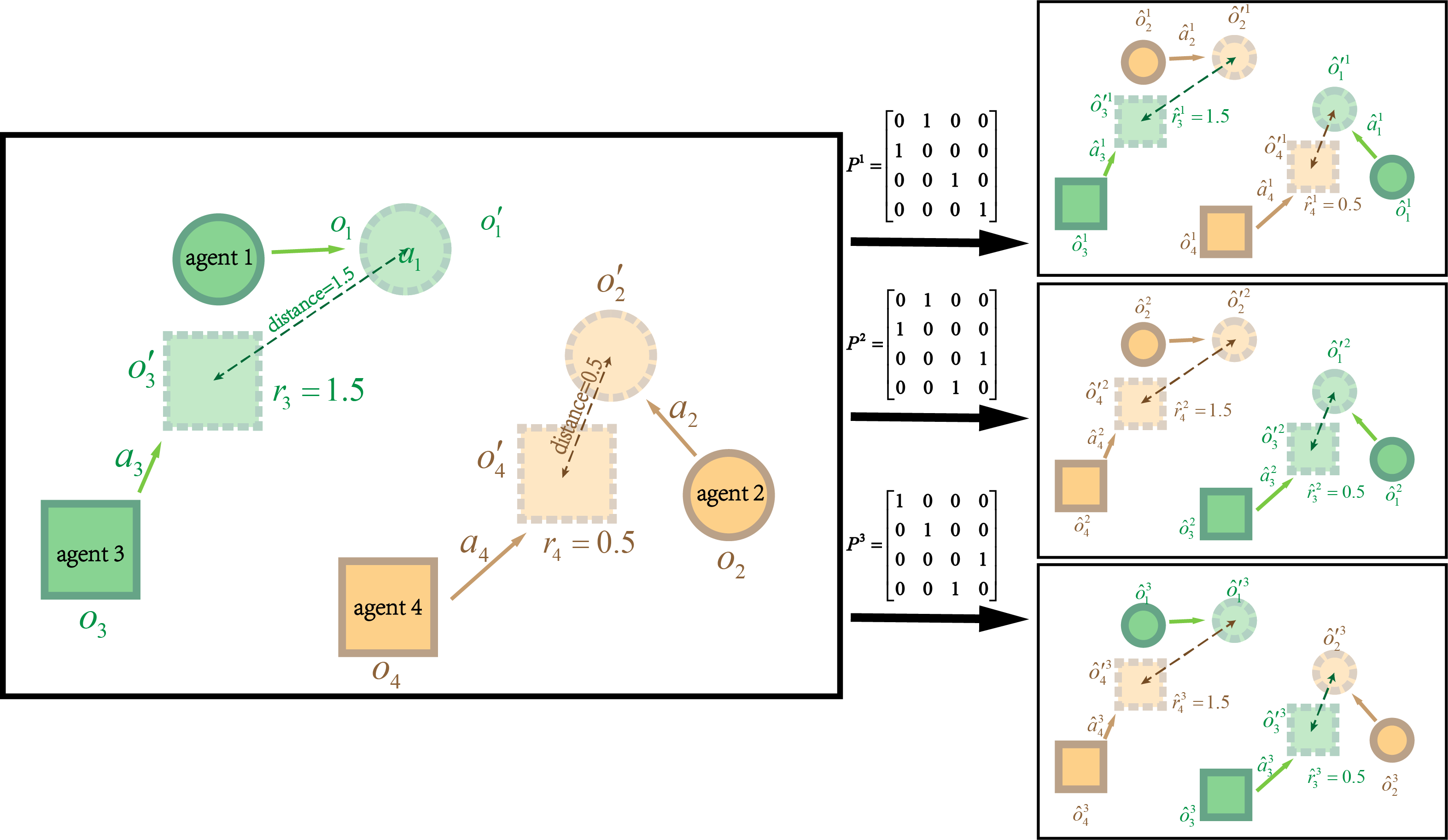}
	\caption{a motivating example to augment the original experience with Shuffle Trick,i.e, shuffle the order of agents (cycles) and adversaries (squares). The shuffling process is done by premultiplying a premutation matrix $P^k$ to the original experience. \textbf{Left:} the original experience;  \textbf{Right:} the generated experiences.
	}
	\label{fig_motivationg_example}
	
\end{figure}

Consider an environment with 4 agents, including 2 good agents (indexed as 1 and 2) and 2 adversaries (indexed as 3 and 4).  The adversaries are rewarded by shortening the distance to its nearest agent. As show in Fig.\ref{fig_motivationg_example}, by shuffling the order of good agents and adversaries respectively, we generate $3$ experiences from the original experience.

Practically, to train a agent $i$, the rewards of other agents are useless. The reward of agent  $i$ in the generated experiences is obtained by choosing the $i^{th}$ element in the generated reward vector $P^k\cdot\vec{r}$.

\subsection{Experience Augmentation}
\label{EA-MADDPG}

In this section,  we consider the concrete method to combine \emph{Shuffle Trick} with MARL algorithms to accelerate training and provide better exploration. To this end, we choose MADDPG\cite{MADDPG2017}, a classical MARL baseline, and analyze the factors that affect its training efficiency and performance.

In recent years, previous work have attempted to improve experience replay in multiple aspects\cite{PER2015}\cite{HER2017}\cite{SHER2020}\cite{MAER2017}. Different from the previous work, we notice that there is another factor that is significantly relevant to the performance of experience replay, i.e., the hyper-parameters named \emph{training interval} $\tau$ and \emph{update times} $n$. As is illustrated in Sec.\ref{preliminaries}, a typical ER-based RL algorithm would update the network parameters for $n$ times when every $\tau$ transitions are added to the replay buffer. An intuitive idea is that increasing the ratio of update times and training interval, $n/\tau$, would result in a improvement of training speed, as it increases the number of updates per unit time.

However, by our experiments, the convergence might suffer when simply enlarging the update times $n$ or decreasing training interval $\tau$ (demonstrated in Sec.\ref{ablation_shuffle}). This phenomenon is possibly caused by the fact that, the replay buffer is an inaccurate subset of the ground truth reward distribution, as the scale of exploration space is far greater than the size of replay buffer, especially in the multi-agent domain. When the ratio $n/\tau$ is tuned to a higher value, the revisited times of every experience in the replay buffer increases proportionally, which may result in the consequnce that the model has been trained so many times on this subset, but not robust enough to encounter the data that don't appear in the replay buffer. In other words, the model has become over-fitted to the dataset in replay buffer when increasing the $n/\tau$ in vanilla MADDPG.

Given the \emph{Shuffle Trick} which provides a fast, thorough and symmetric exploration of the observation action space, we notice that extra update times on the generated dataset could effectively accelerate the training without deteriorating the convergence; on the contrary, in some environments it boosts the result by a significant margin. We refer this training technique to \emph{Experience Augmentation}. There are two steps to perform experience augmentation in off-policy MARL: firstly, use shuffle trick to generate $E$ extra experiences; then, update parameters on these experiences sequentially, where $E$ is a newly introduced hyper-parameter called \emph{EA-times} ,which is a shorthand of \emph{Experience Augmentation extra update Times}. This hyper-parameter determines how many times the model additionally train on the generated dataset, whose effected are examined in Sec.\ref{ablation_eatimes}. 

\begin{figure}
	\centering
	\includegraphics[width=12cm]{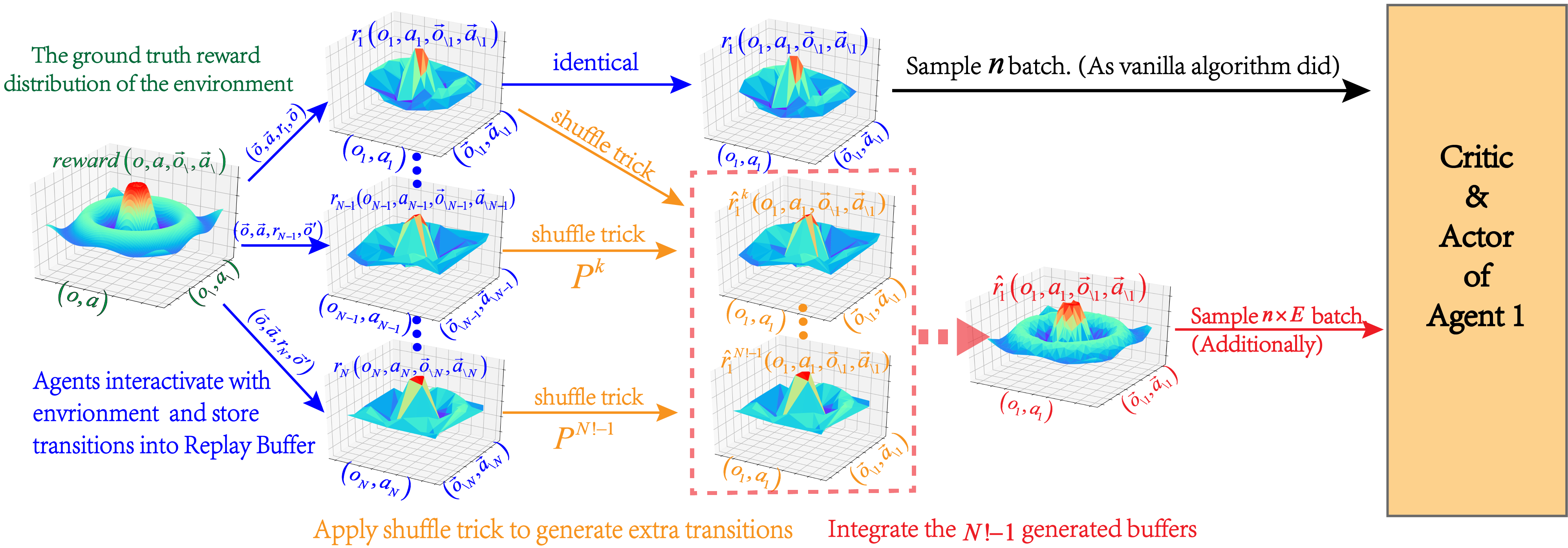}
	\caption{
		experience augmentation for a cooperative envrionment with $N$ homogeneous agents, training agent $1$. \textbf{Green:} the ground truth observation-action-reward distribution of the envrionment. \textbf{Blue:} sample experiences $(\vec{o},\vec{a},\vec{r},\vec{o'})$ into the memory buffer \textbf{Black:} randomly sample experience from the buffer that stores $(\vec{o},\vec{a},r_1,\vec{o'})$, every experience will be revisited for $batch\_size\times\frac{ n}{\tau}$ times in average.\textbf{Brown:} Use shuffle trick to generate $N!-1$ generated replay buffer for agent $1$. \textbf{Red:} Integrated the $N!-1$ generated buffers into a thorougher, smoother one, then sample  $E$ experiences from this buffer at a time, every experience will be revisited for $batch\_size\times\frac{ E}{\tau\times (N!-1)}$ times in average.
	}
	\label{figure_EA-MADDPG}
\end{figure}

The principle behind the effectiveness of EA and the whole process of experience augmentation, are shown in Fig.\ref{figure_EA-MADDPG}. It can be seen that in vanilla MARL, the parameters only update on every experience in the original buffer(which is a coarse approximation to groud truth reward distribution) for multiple times; while with EA, the parameters are additionally trained on the integrated buffer (which is a far more accurate approximation to groud truth) for \emph{EA-times}. This new feature allows the Q-function to provide more accurate estimation of Q-value and helps the policy to find better local minimum.

\section{Experiments}
The experiments part is organized as follows. In Sec.\ref{environments} we introduce MARL environments we use for the experiments. In Sec.\ref{experiemntal_settings} we present the experimental settings. In Sec.\ref{performance} we compare the performance of MADDPG with and without EA and analyze the accelerating effect of EA in the training process in realistic time. In Sec.\ref{ablation} we do ablation studies to prove the necessity of shuffle trick, and examine the effect of the newly introduced hyper-parameter EA-times. 
\subsection{Envrionments}
\label{environments}
Our experiments are mainly based on MPE\cite{MADDPG2017}\cite{emergence2018}. We consider the following three tasks: Cooperative Navigation, UAV used for Mobile Base Station, World with Communication. The first two are homogeneous environments, and the last one is a heterogeneous environment with two groups of agents. The details of the envrionments are as follow.
\paragraph{Cooperative Navigation}
Cooperative navigation consists of $N$ agents and $L$ landmarks. The goal of this environment is to occupy all of the landmarks while avoiding collisions among agents. By training, the agents learned the assignment strategy to cover the landmarks. Each agent receives a negative reward $R_1$, which is the negative of the sum of the distances between the L landmarks to their nearest agents, and is shared by all agent; it would also receive a negative reward  $R_2$ if it collides to another agent. In this environment, we set $R_2 = -2$, and simulate one case: $N=3,L=3$
\paragraph{UAV used for Mobile Base Station}
The scenario where UAV swarm are used for Mobile Base Station (MBS), was proposed by \cite{uav2019}, and implemented by us using MPE\cite{MADDPG2017}. The environment consists of $N$ UAVs and $L$ PoIs, where UAVs work as mobile base stations to provide communication services to the public (abstracted as to cover a set of PoI). Note that PoIs are invisible to UAVs. The reward was set to encourage the UAVs to cover more PoIs, meanwhile, it takes into account the fairness among the covered time for every PoI and the efficiency of energy consumption. By training, the UAVs learned the \textbf{latent distribution} of PoIs and the corresponding moving strategies. In this environment, we simulate one case: $N=3,L=25$ 
\paragraph{World with Communication}
\emph{World with Communication} consists of $N$ slower predators work together to chase $M$ fast-moving preys, and $L$ inaccessible obstacle, as well as $F$ accessible forest. The observation of each agent is the concatenation of its position and velocity, the locations of obstacle, the locations of forest, and the locations of other agents. One of the N predators is the leader who can see preys hiding in the forest and can share the information with other predators through the communication channel. The predators get positive reward $R_1$ when colliding with any preys. Preys get negative reward $R_2$ when caught by (colliding with) any predators. In this environment, we set $R_1=10$,$R_2=-10$, $R_3=2$,and simulate one case:$ N=4$,$M=2$,$L=1$,$F=1$.
\subsection{Experienmental Setup}
\label{experiemntal_settings}
Following MADDPG, the actor policy and critic are both parameterized by a two-layer MLP with 128 hidden units per layer and ReLU activation function. Adam is used as the optimizer. The size of the replay buffer is one million. The batch size is 1024. The discounted factor is 0.95. The training interval $\tau$ is 100 and update times $n$ is 1, as suggested by MADDPG. The learning rate for each environment is decided by a coarse grid search. For UAV Mobile Base Station, the learning rate is fixed to 0.01.  For other scenarios, the learning rates are fixed to 0.001. For consistency, we set EA-time $E$ to 3 in every environment. In the ablation studies, we examine the impact of EA-times, the cooperative navigation scenario was chosen, and $E\in\{0,1,3,5,7,15,31\}$ was tested.

For each case, we tried at least 10 random seeds. We train our models until convergence (either 80,000 or 160,000 episodes), and then evaluate them by averaging the metrics in the last 20,000 episodes. The mean value(solid line in figure) and quantile(translucent part in figure) are analyzed.
\subsection{The performance and time-efficiency of EA}
\label{performance}

\paragraph{Does EA boost the convergence?}
In order to verify if EA improves performance we evaluate MADDPG with and without EA on all 4 tasks. Moreover, we compare MADDPG with EA (EA-MADDPG) against MADDPG with PER (PER-MADDPG), and the performance of DDPG is also tested. For EA-MADDPG, during each update, it will generate 3 extra experiences and upate the parameters on them sequentially.

We present the results in two homogeneous environments in Table.\ref{table_spread} and Table.\ref{table_uav}. It is clear that EA-MADDPG outperforms other algorithms by a significant margin. To evaluate the approach in heterogeneous environments, we pitch EA-MADDPG agents against MADDPG agents and DDPG agents, and compare the result of the agents and adversaries in Table. \ref{table_world}. It shows that agents with EA take more advantages over their opponents without EA.

\begin{figure}
	
	\centering
	\subfigure[cooperative navigation]{
		\label{figure_time_a}
		\includegraphics[width=4cm]{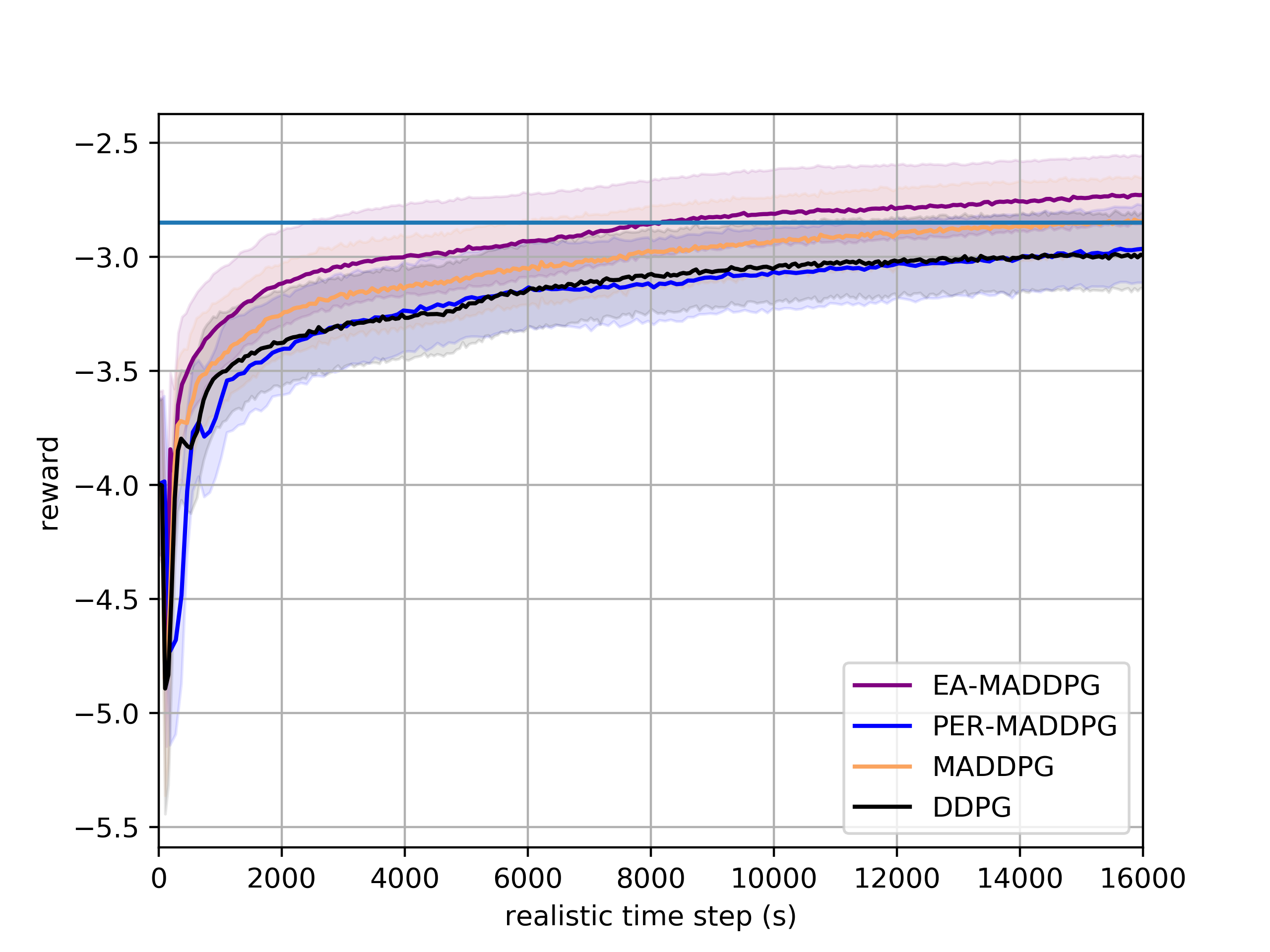}}
	\subfigure[UAV for MBS]{
		\label{figure_time_b}
		\includegraphics[width=4cm]{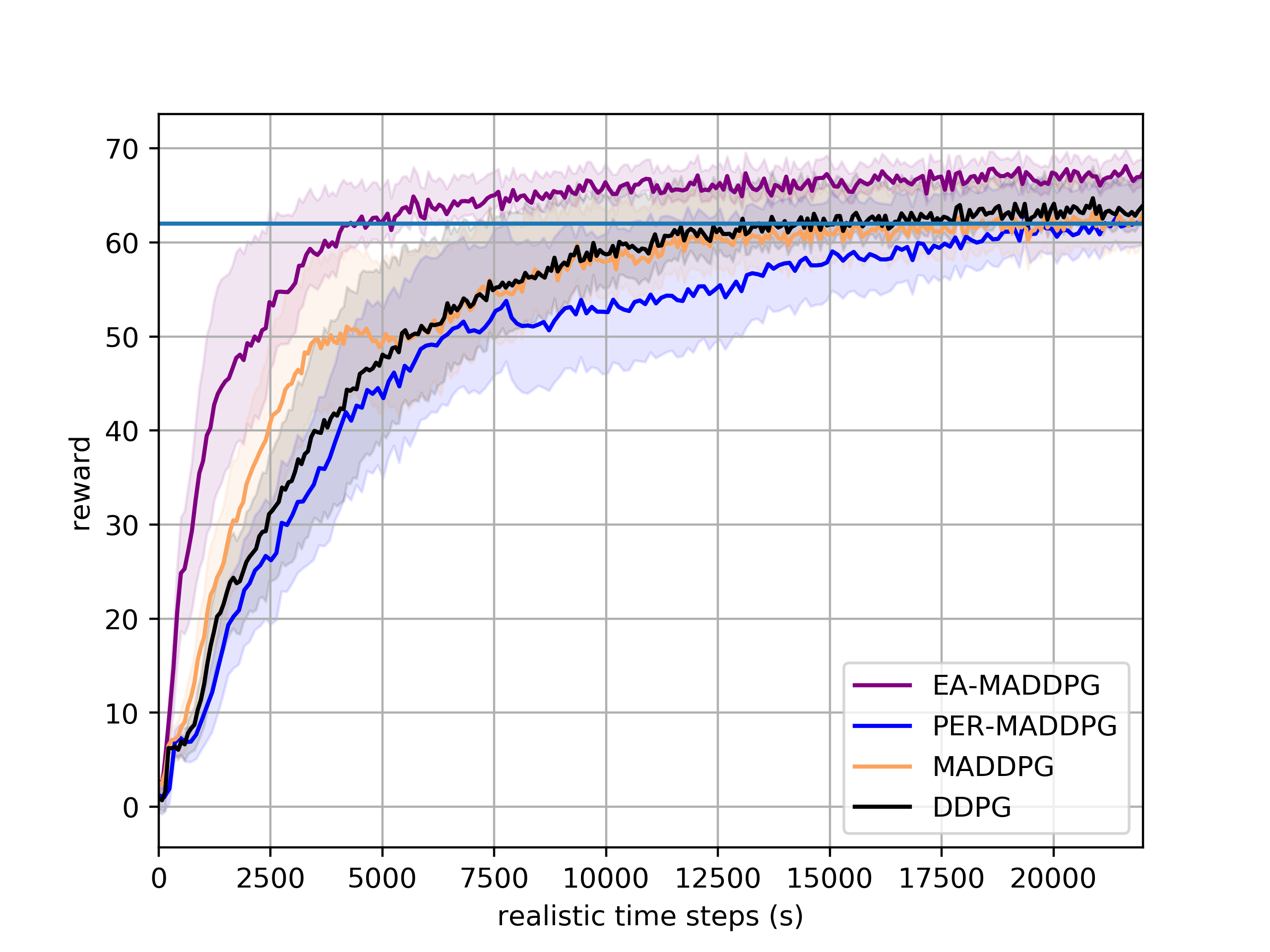}}
	\subfigure[world bad]{
		\label{figure_time_c}
		\includegraphics[width=4.5cm]{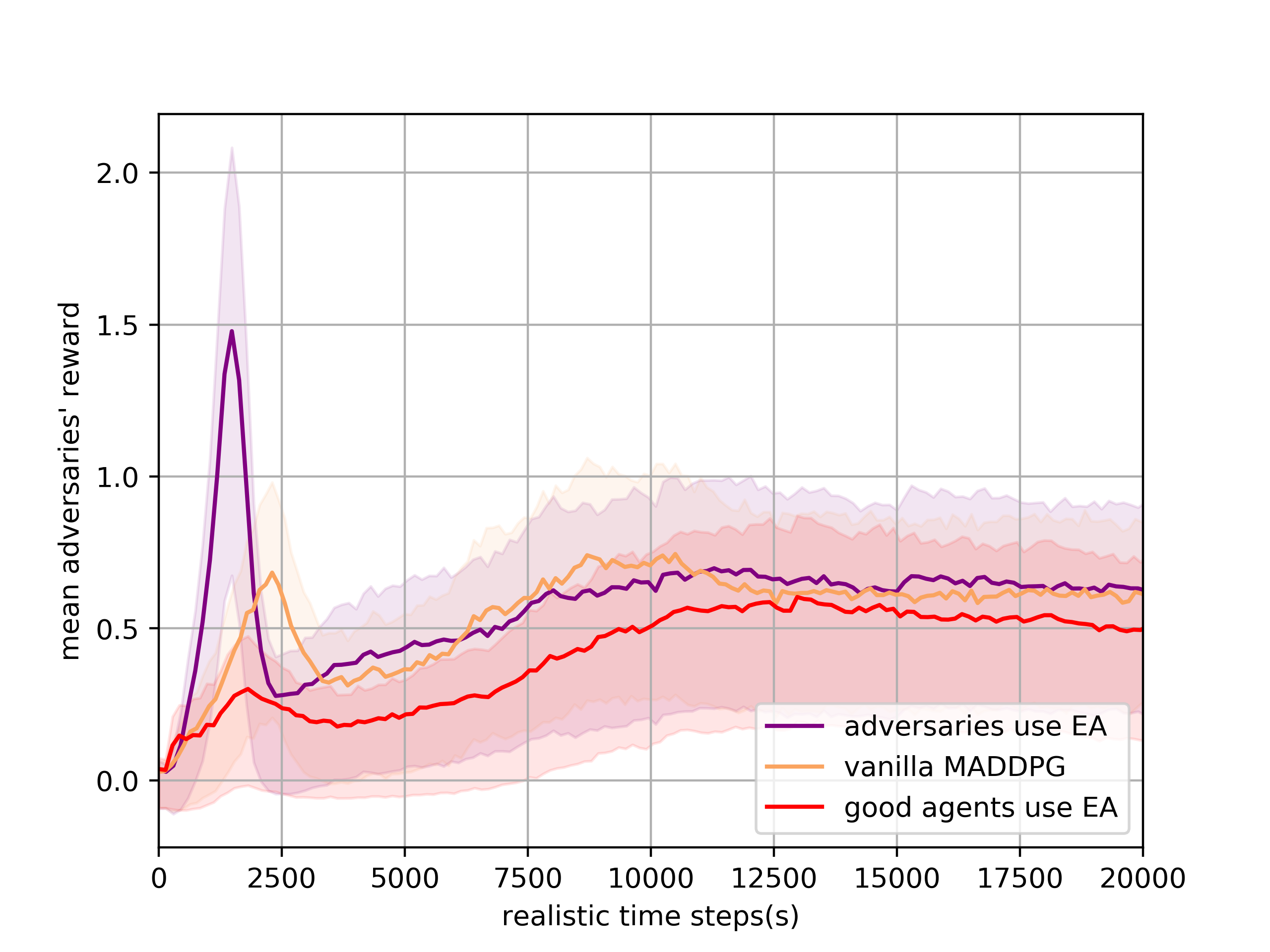}}
	\caption{
		The learning curves of cooperative navigation and UAV for MBS. (Note that the relatively slow training speed of PER is partially due to the extra computational complexity $O(logn)$ for each importance sampling, and our implementation of PER is not properly optimized.)
	}
	\label{figure_time}
\end{figure}

\paragraph{Does EA accelerate the training?}
\label{acceleration}
Given the fact that EA improves the convergence of vanilla MADDPG, the next problem is that whether EA could accelerate the training process of MADDPG. From Fig.\ref{figure_time_a}\ref{figure_time_b}, which present the learning curves in two homogeneous environments, it is clear that the training of EA-MADDPG is much faster than vanilla MADDPG and other algorithms, in terms of realistic training time. Note that in UAV scenario, EA-MADDPG achieves the reward equal to the convergence result of vanilla MADDPG with only 1/4 time. EA also accelerates the training of heterogeneous tasks in the early stage. As shown in Fig.\ref{figure_time_c}, the agents with EA take the advatange over their opponents earlier than those without EA (see the first peak in the learning curve).
\begin{table}
	\caption{result of different algorithms in Cooperative Navigation environment}
	\label{table_spread}
	\centering
	\begin{tabular}{lllll}
		\toprule
		Algorithm     & distance  & collision & reward & time(s/1000 episode) \\
		\midrule
		EA-MADDPG(ours) & 0.2256 & 76.3785 & -2.0735 & 126.7430\\
		PER-MADDPG & 0.2846&77.3508&-2.8815& 185.9162\\
		MADDPG    &0.2448& 77.4650&-2.8221& 102.9587\\
		DDPG  & 0.3089&76.9333&-2.9952&104.2302\\
		\bottomrule
	\end{tabular}
\end{table}
\begin{table}
	\caption{result of different algorithms in UAV for MBS environment}
	\label{table_uav}
	\centering
	\begin{tabular}{llllll}
		\toprule
		Algorithm     & coverage     & fairness & efficiency & reward & time(s/1000 episode) \\
		\midrule
		EA-MADDPG(ours) & 16.4007  &0.6250      &1.4498  &  67.1151&165.8283\\
		PER-MADDPG & 16.0238&0.6096&1.4576&63.7209&239.5142\\
		MADDPG    &15.8808&0.6038&1.5134&62.2762&145.9051\\
		DDPG  & 15.9035&0.6053&1.4263&63.4254&145.5347\\
		\bottomrule
	\end{tabular}
\end{table}
\begin{table}
	\caption{result of different algorithms in World with Communication environment}
	\label{table_world}
	\centering
	\begin{tabular}{llllll}
		\toprule
		\multicolumn{2}{c}{Algorithm}                   \\
		\cmidrule(r){1-2}
		agent     & adversary   & caught & agent reward & adversary reward  & training time \\
		\midrule
		MADDPG & MADDPG & 3.2495&-0.3348& 0.6097 &256.726\\
		MADDPG & EA-MADDPG & 3.3295& -0.3334 & 0.6214 & 297.721\\
		EA-MADDPG  & MADDPG & 2.7668 &-0.2810 & 0.5050 & 279.427 \\
		MADDPG  & DDPG & 3.1528 &-0.3318 & 0.5869 & 259.038\\
		DDPG & MADDPG & 2.8757 & -0.2957  & 0.5176 & 259.154\\
		\bottomrule
	\end{tabular}
\end{table}

\subsection{Ablation Studies}
\label{ablation}
\begin{figure}
	
	\centering
	\subfigure[the necessity of shuffle trick]{
		\label{figure_ablation_shuffle}
		\includegraphics[width=4.5cm]{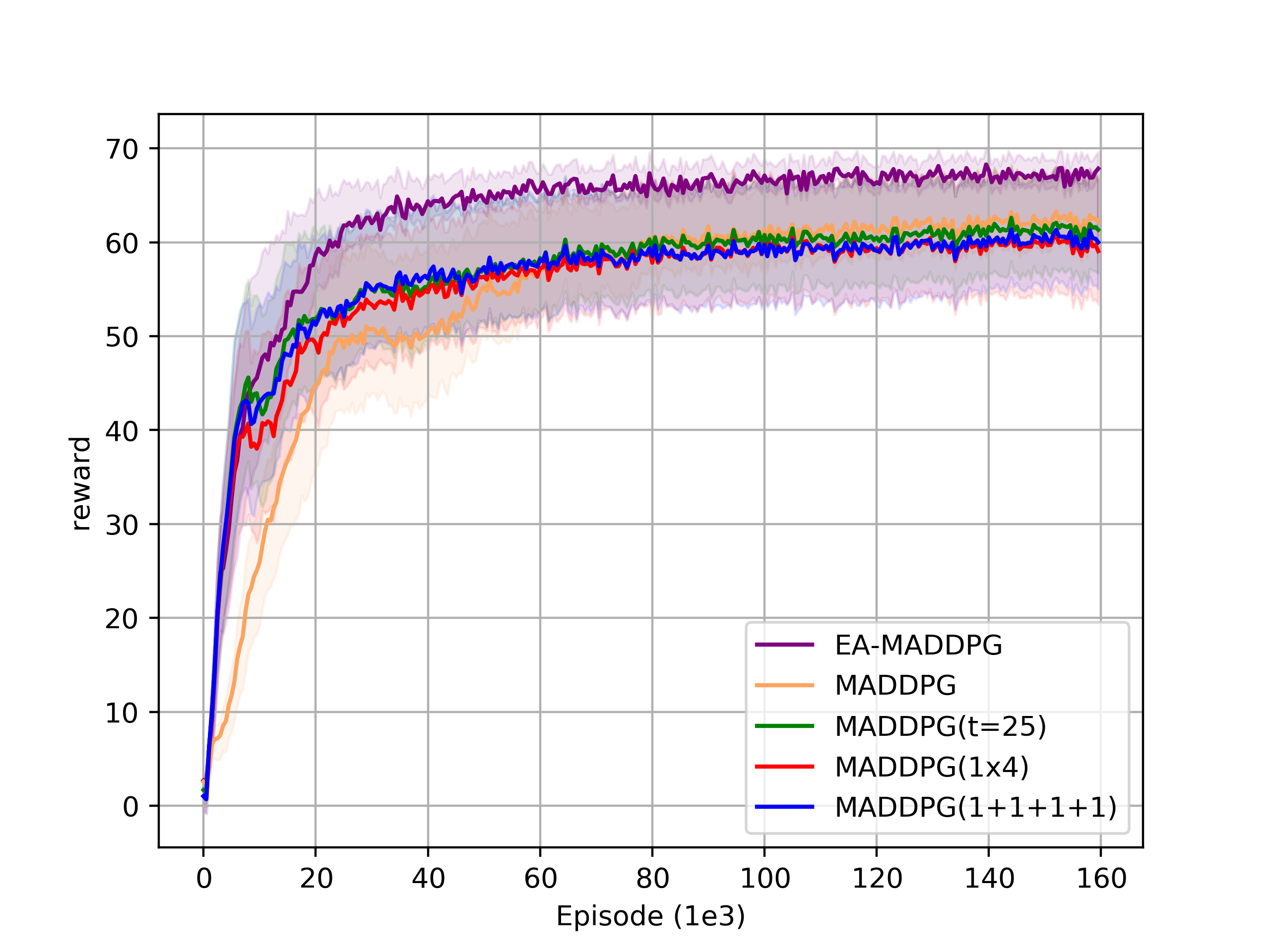}}
	\subfigure[the effect of EA-times]{
		\label{figure_ablation_eatimes}
		\includegraphics[width=4.5cm]{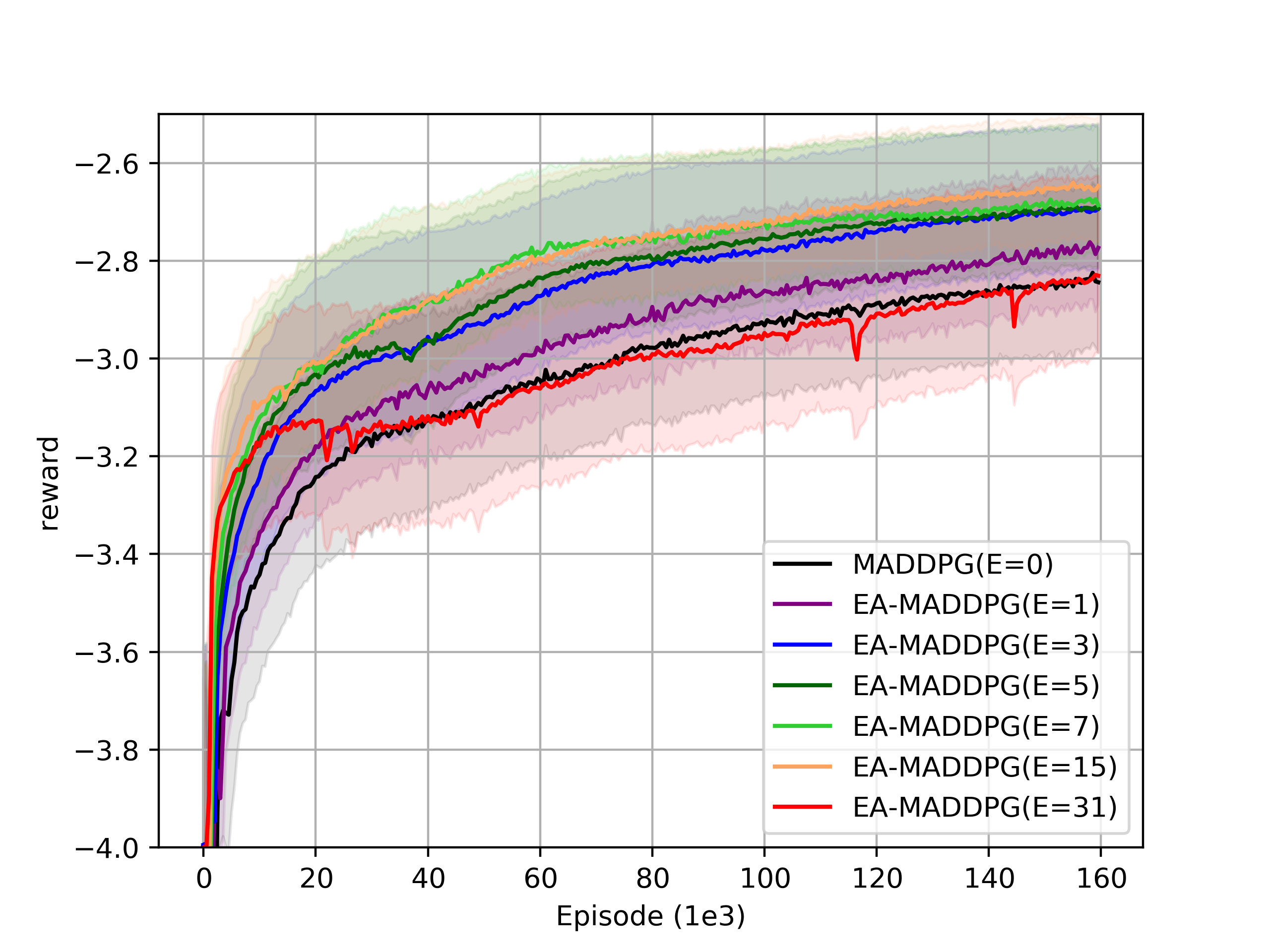}}
	\subfigure[the effect of EA-times]{
		\label{figure_ablation_eatimes_collision}
		\includegraphics[width=4.5cm]{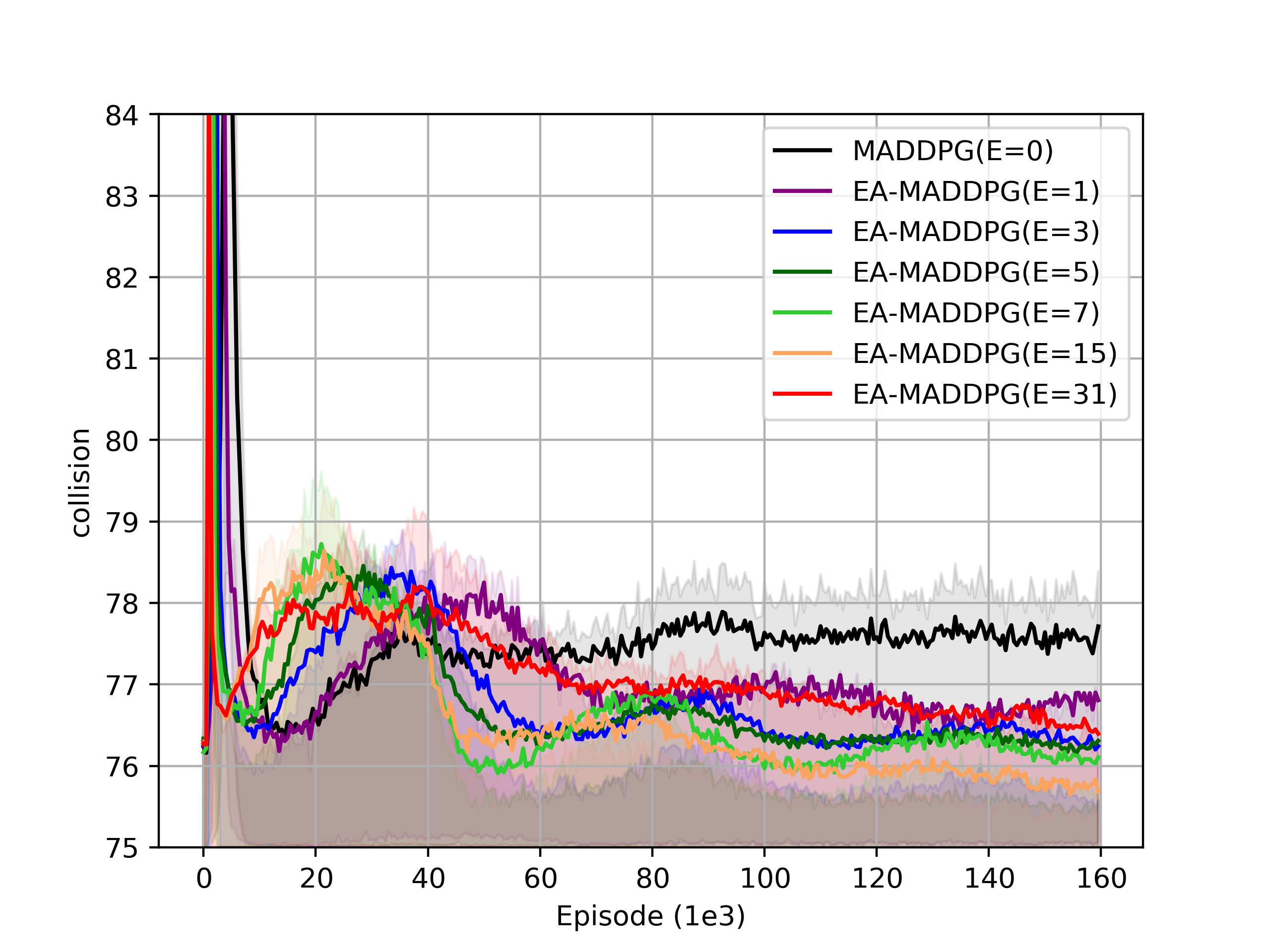}}
	\caption{
		the results of ablation studies.\textbf{Left:} learning curves of EA-MADDPG and other 3 training strategies which has the same $n/\tau$ in the UAV for MBS environment. \textbf{Middle\&Right}: learning curves of EA-MADDPG with different EA-times in the cooperative navigation environment.
	}
	\label{figure_bar}
\end{figure}

With the verification of the performance of the \emph{Experience Augmentation} technique, we performed in-depth ablation studies regarding the technique with respect to two key aspects of the technique: 1) the \emph{Shuffle Trick}, 2) the \emph{EA-Times}. In this section, we have 2 main purposes: 1. to demonstrate the necessity of Shuffle Trick, the technique we proposed in Sec.\ref{ShuffleTrick} to generated extra experiences; 2. to analyze the function of the EA-times.
\label{ablation_shuffle}
\paragraph{The Necessity of Shuffle Trick}
One may argue that the increased $n/\tau$ caused by EA-times it the key factor of the performance of EA. To demonstrate the necessity of the shuffle trick, we examine 3 cases which have the same $n/\tau$ with EA-MADDPG(EA-times=3), they are: 1.MADDPG(t=25),which updates the network in every 25 iterations; 2.MADDPG(1+1+1+1), which updates the network with 4 sampled batches of experience in every 100 iterations; 3.MADDPG(1x4), which updates the network with a same batch of experience for 4 times in every 100 iterations. The environment is UAV for mobile base station, and the learning rate is fixed to 0.001. At least 10 random seeds were used for each case. As expected, it is seen in Fig.\ref{figure_ablation_shuffle} that the EA-MADDPG(EA-times=3) outperforms vanilla MADDPG in terms of learning speed and convergence result. The MADDPG(t=25) case, MADDPG(1+1+1+1) and MADDPG(1x4), which need close training time for each episode with EA-MADDPG(EA-times=3), show close training speed with EA-MADDPG in the first 10,000 episodes, yet lead to a worse convergence, when  compared with Vanilla MADDPG. A reason for this phenomenon is, possibly, the model has been over-fitted to the replay buffer, as illustrated in Sec.\ref{EA-MADDPG}. 
\paragraph{The Effect of EA-Times}
\label{ablation_eatimes}
To study the effect of EA-times, we compare the performance of EA-MADDPG with different EA-times varying from $\{0,1,3,5,7,15,31\}$.  The environment is Cooperative Navigation, where $N=3$ and $L=3$ . Note that EA-MADDPG with EA-times=0 corresponds to vanilla MADDPG.At least 10 random seeds are used for each case. Table.\ref{table_eatimes} shows the performance and training time for each episode of each case, to help selecting best EA-times for the trade-off between performance and training speed. The learning curves of each case are also shown in Fig.\ref{figure_ablation_eatimes}\ref{figure_ablation_eatimes_collision}. It is seen that EA-MADDPG with any EA-times (except 31) significantly outperforms vanilla MADDPG. As expected, the learning speed (in terms of episode) increases as the EA-times increases. It can also be seen that in a suitable range (less than 31), as EA-time increases, the convergence result becomes better, and the boost in both performance and training speed slows down as EA-time exceeds 3. Considering the training speed in realistic time, we suggest that the most efficient value of EA-times in this envrionment is within the range of 3 to 7.

\begin{table}
	\caption{result of different EA-times in Cooperative Navigation}
	\label{table_eatimes}
	\centering
	\begin{tabular}{lllll}
		\toprule
		Algorithm     & distance  & collision & reward & training time \\
		\midrule
		MADDPG(E=0) & 0.2550 & 77.5718 & -2.8505 & 102.959\\
		EA-MADDPG(E=1) & 0.2468 & 76.7440 & -2.7858 & 109.051\\
		EA-MADDPG(E=3) & 0.2256 & 76.3785 & -2.7035 & 126.743\\
		EA-MADDPG(E=5) & 0.2216 & 76.2755 & -2.6979 & 137.503\\
		EA-MADDPG(E=7) & 0.2154 & 76.1493 & -2.6859 & 158.636\\
		EA-MADDPG(E=15) & 0.2049 & 75.8045 & -2.6557 & 202.314\\
		EA-MADDPG(E=31) & 0.2728 & 76.5594 & -2.8543 & 299.762\\
		\bottomrule
	\end{tabular}
\end{table}

\section{Related Work}
\emph{Experience Replay}\cite{lin1992} has became a fundamental component of off-policy RL after it was used in DQN\cite{DQN2015}. Previous work have attempted to improve experience replay in multiple aspects, such as \emph{Prioritized Experience Replay} \cite{PER2015}, which prioritizes experiences in the replay buffer to speed up training, and \emph{Hindsight Experience Replay} \cite{HER2017}, which addresses the sparse reward problem by introducing a sub-goal $g$ and giving reward of the transition based on the sub-goal.  There are also some work focused on the multi-agent domain, e.g. \cite{MAER2017} using a multi-agent variant of importance sampling to naturally decay obsolete data and conditioning each agent’s value function on a fingerprint that disambiguates the age of the data sampled from the replay memory. All of them are orthogonal to our work and can be easily combined to get further improvement.

Our approach, especially the \emph{Shuffle Trick}, may be seen as a novel form of exploration strategy in multi-agent domain. Simple exploration strategies such as $\epsilon-greedy$ , which is used in this paper and MADDPG, may need exponentially many steps to find a (near-)optimal policy \cite{complexity1991}. By using shuffle trick, we could factorially augment the original dataset, which is a big acceleration of exploration. There are also many exploration strategies successfully applied to deep reinforcement learning and demonstrated their performance, such as \cite{unifying2016}, \cite{exploration2017}, \cite{explicit2019}. Both of them are orthogonal to our work and can be combined for faster exploration in multi-agent domain.

Interestingly, a data augmentation method used in \cite{PIC2019} is closed to our method. They shuffles the order of agents' observations and actions (they assume that every agent in a group share the reward value), and then let the parameters only trained on the generated experience. The difference between \cite{PIC2019} and ours is that, firstly, our \emph{Shuffle Trick} is applicable to the homogeneous and heterogeneous environment; secondly, we train on the original experience, then train on the generated experiences for \emph{EA-times}, making full use of the generated dataset and accelerates the training.
\section{Conclusions}

In this paper, we present a novel technique called \emph{Shuffle Trick} to perform a fast, thorough and symmetric exploration that factorially expands the original dataset in multi-agent domain. Based on the shuffle trick, we introduce a time-efficient training method called \emph{Experience Augmentation}, which accelerates the training and boosts the convergence in off-policy MARL. We experimentally demonstrate that with MADDPG in two homogeneous environments and one heterogeneous environment. In addition, we have carried out in-depth ablation studies on the proposed algorithm, proved the necessity of shuffle trick, and examined the effects of EA-times on training speed and convergence. 

\bibliographystyle{plainnat}
\bibliography{ref} 
\end{document}